\RequirePackage{silence}
\documentclass{article}

\usepackage{microtype}
\usepackage{graphicx}
\usepackage{subcaption}
\usepackage{booktabs}
\usepackage{float}
\usepackage[hyperfootnotes=false]{hyperref}

\usepackage[accepted]{icml2026}
\usepackage{amsmath}
\usepackage{amssymb}
\usepackage{mathtools}
\usepackage{amsthm}
\usepackage[capitalize,noabbrev]{cleveref}
\usepackage[disable,textsize=tiny]{todonotes}

\theoremstyle{plain}

\theoremstyle{definition}

\theoremstyle{remark}

\newcommand{\model}{VideoLLaMA 2-7B-AV}
\newcommand{\snap}{\ell^*}
\newcommand{\avproxy}{s_{AV}^{(\ell)}}

\icmltitlerunning{Compositional Failure in Audio-Visual LLMs: Late-Layer Prior Dominance Under Cross-modal Conflict}

\begin{document}

\twocolumn[
  \icmltitle{Compositional Failure in Audio-Visual LLMs:\texorpdfstring{\\}{ } Late-Layer Prior Dominance Under Cross-modal Conflict}

\icmlsetsymbol{equal}{*}
  \icmlsetsymbol{senior}{$\dagger$} 

\begin{icmlauthorlist}
    \icmlauthor{Adarsh Sudheer}{equal}
    \icmlauthor{David Li}{equal}
    \icmlauthor{Omar Elbanna}{}
    \icmlauthor{Ishaan Kodarapu}{}
    \icmlauthor{Arjun Bahuguna}{senior}
    \icmlauthor{Vasu Sharma}{senior}
\end{icmlauthorlist}

  \icmlcorrespondingauthor{Adarsh Sudheer}{adarshsudheer09@gmail.com}
  \icmlcorrespondingauthor{David Li}{davidli07712@gmail.com}

  \icmlkeywords{Mechanistic Interpretability, Audio-Visual LLMs, Multi-modal Reasoning}

  \vskip 0.3in
]

\printAffiliationsAndNotice{* Equal contribution. \quad
  $\dagger$ Senior author.}

\begin{abstract}
We study audio-visual conflict as a compositional
generalization test for AV-LLMs: the model must
combine synchronized but semantically incompatible audio and video evidence and decide whether the pair matches. On VideoLLaMA 2-7B-AV, three alignment configurations remain nearchance on the scored exact-string Yes/No  subset of AVHBench, even though their output priors
shift substantially. Similarly, off-the-shelf InternVideo2 experienced a 32.3$\%$ accuracy decrease
specifically under cross-modal conflict, accompanied by a 17.3$\%$ instruction-following failure. We
call this failure mode \textit{prior dominance}: late-layer commitment to an internally preferred answer pattern that is weakly grounded in the conflicting
inputs. To explain this behavior, we conduct a
mechanistic interpretability analysis and find that
commitment remains concentrated at $25.5 \pm 1$ layers. We show that stronger temporal alignment
changes answer bias, but do not improve compositional conflict resolution. Code and data to reproduce our mechanistic audit and behavioral evaluations are available at \url{https://github.com/AdarshSudheer09/AVHBench-dmai}.
\end{abstract}

\section{Introduction}
Audio-visual large language models are increasingly evaluated on benchmarks for grounding, hallucination, and synchronized understanding
\cite{wang2024internvideo2,cheng2024videollama2,sungbin2025avhbench,jung2025avcd,leng2024cmm}.  In our study the key question is: can a model \textit{compose} evidence across modalities when the two streams conflict? If a video depicts a barking dog while the audio
contains a revving engine, each modality is locally plausible, but the pair is jointly inconsistent. Solving this case
requires cross-modal composition rather than a uni-modal
shortcut.

This conflict setting is a useful stress test because agreement
examples are often solvable with shallow co-occurrence on
a single modality, whereas conflict examples reveal whether
the model compares streams or defaults to a high-probability
prior. The prevailing assumption that greater multi-modal
alignment improves reasoning has driven work in contrastive
pre-training, temporal synchronization, and token-level fusion. However, our findings challenge this assumption, implying that alignment-stage interventions change answer
bias without recovering compositional conflict resolution,
which is consistent with a pattern where learned prior dominates generation before cross-modal comparison can occur.
In our experiments with VideoLLaMA 2-7B-AV \cite{cheng2024videollama2}, increasingly strong forced alignment methods like Audio-Conditioned Token Concatenation (ACTC),
Timestamp-Aware Token Interleaving (TATI), and Asymmetric Modality Dropout (AMD) change answer bias, but
do not improve compositional conflict resolution. To establish that this vulnerability extends beyond our alignment
pipeline, we additionally use an off-the-shelf model, InternVideo2 \cite{wang2024internvideo2}. On the full AVHBench dataset,
while the model achieves $60.1\%$ accuracy, the performance
drops to $27.8\%$ under contradiction, a $32.3\%$ decrease. Because performance drops significantly below $50\%$ on a binary classification task, the model is not guessing, but is
systematically misled.

We use \textbf{prior dominance} as an explanatory hypothesis for
this regime: the model's final solution is increasingly explained by an internally preferred response pattern rather
than by a composition of the auditory and visual evidence, a
claim that we support with logit-lens mechanistic evidence.
The contributions of this paper are as follows. First, it reframes audio-visual conflict as a compositional learning
problem rather than only a hallucination benchmark issue.
Second, it demonstrates that stronger temporal alignment
can change answer bias without improving conflict set accuracy. Third, it localizes a stable late-layer commitment
point in the model using logit-lens.

\section{Conflict-Based Evaluation and Notation}
Let $X_v$ and $X_a$ denote the video and audio streams. A conflict example is one in which each stream is individually coherent but the pair is semantically incompatible. In this
paper, compositional success requires four ordered steps:
extract audio evidence from $X_a$, extract video evidence
from $X_v$, temporarily bind both streams, and finally judge
their compatibility–whether the two streams jointly describe
the same event.

We probe where the model's answer becomes committed using a logit-lens operator,
{\setlength{\abovedisplayskip}{5pt}%
\setlength{\belowdisplayskip}{5pt}%
\[
\hat{p}^{(\ell)} = \mathrm{softmax}\!\bigl(W_U\,\mathrm{RMSNorm}(h^{(\ell)})\bigr),
\]
where $h^{(\ell)}$ is the residual stream at layer $\ell$ and $W_U$ is the final unembedding matrix. We define the snap layer \(\snap\)
as the earliest layer after which the top prediction remains
unchanged through the final layer. We use \(\snap\)
as a behavioral
probe for prior dominance, with an early stable commitment
that suggests the model has decided on an answer before
fully processing cross-modal evidence.

\section{Methodology}

We investigate compositional failure by evaluating both
plain models and fine-tuned models. We utilize InternVideo2 \cite{wang2024internvideo2} to establish the baseline behavioral collapse under cross-modal conflict. We then conduct
an in-depth behavioral and mechanistic audit using VideoLLaMA 2-7B-AV \cite{cheng2024videollama2}, applying three distinct
alignment configurations to test whether post-training interventions resolve the deficit.

\subsection{Evaluation Dataset}
Standard benchmarks reward modality agreement. We curated an adversarial conflict split (N = 1,281) from AVCD
and AVHBench using automated filtering to isolate severe
semantic contradictions (e.g., a dog visual paired with car audio). We measure the influence of various modality grounding methods on compositional conflict-resolution using this
filtered split.

Crucially, this filtering inadvertently introduced a severe
label imbalance: $92\%$ of the resulting ground truths were
”No.” We identify this as a critical methodological trap.
Evaluating on such skewed adversarial splits allows models to achieve illusory accuracy gains (e.g., $70\%$) simply
by adopting a rejection bias. To prevent these statistical
illusions, our final pipeline evaluation relies strictly on the
full AVHBench dataset (N $\approx$ 6,300). This maintains a true
50/50 class balance, ensuring chance-level performance is
actually $50\%$ and forcing the model to demonstrate genuine
multimodal reasoning.

\subsection{Three-Stage Alignment Pipeline}

We investigate the effects of a three-stage fine-tuning pipeline on 
VideoLLaMA2-7B, designed to progressively increase the degree of explicit 
audio-visual grounding. Each stage is evaluated independently to isolate its 
contribution to model behavior. All pipeline stages were implemented using LoRA on the frozen VideoLLaMA2-7B-AV base model; full training and hardware configurations are detailed in Appendix B.

\subsubsection{Audio-Conditioned Token Concatenation (ACTC)}

ACTC serves as our baseline integration stage. Audio features extracted from 
BEATs are projected into a shared 3,584-dimensional embedding space via an 
STCConnector, and then concatenated sequentially with the visual token prefix 
before being passed to the language model. While this gives the model access 
to both modalities, the audio and visual tokens remain temporally unaligned 
within the sequence, leaving the model free to attend to either modality 
independently without explicit cross-modal synchronization.

\subsubsection{Timestamp-Aware Token Interleaving (TATI)}

TATI introduces explicit temporal synchronization by replacing sequential 
concatenation with a synchronous 1:1 interleaving of visual and audio tokens adapted from the AVTI mechanism introduced by AVicuna \cite{tang2024avicuna}. 
For each of the 16 sampled video frames, the corresponding audio token is 
placed immediately adjacent in the sequence. A shared learnable temporal 
embedding $E_{\text{temp}}$ is added to both modalities at each timestep $t$:
{\setlength{\abovedisplayskip}{6pt}%
\setlength{\belowdisplayskip}{6pt}%
\begin{equation}
    \tilde{v}_t = v_t + E_{\text{temp}}(t), \quad \tilde{a}_t = a_t + 
    E_{\text{temp}}(t)
\end{equation}
This forces the model to process each visual frame alongside its temporally 
corresponding audio token, with the intention of reducing temporal confusion 
during multimodal reasoning. 

\subsubsection{Asymmetric Modality Dropout (AMD)}

AMD is applied as an attempted regularizer during supervised fine-tuning. At each training step, a stochastic dropout
mask is applied to either the visual or audio token sequence
with probability $P_{\text{mask}}$, effectively blinding the model to one
modality at a time. While the goal is to stop the model from
developing a lazy reliance on a single modality, the true
effect is uncertain. Modality dropout risks unintentionally
teaching the model that one modality is enough, which can
encourage bypassing the contradiction checks.

\subsection{Evaluation Protocol}

InternVideo2 and each ablation checkpoint were evaluated
on the full AVHBench dataset (N $\approx$ 6,300) as well as the
curated 1,281-sample conflict split. All evaluations used greedy decoding with a temperature of 0.01. Yes/No accuracy was computed via exact string match, while captioning
outputs were analyzed qualitatively given the known failure
of exact match metrics for open-ended generation.

\begin{figure}[t]
\centering
\includegraphics[width=\columnwidth]{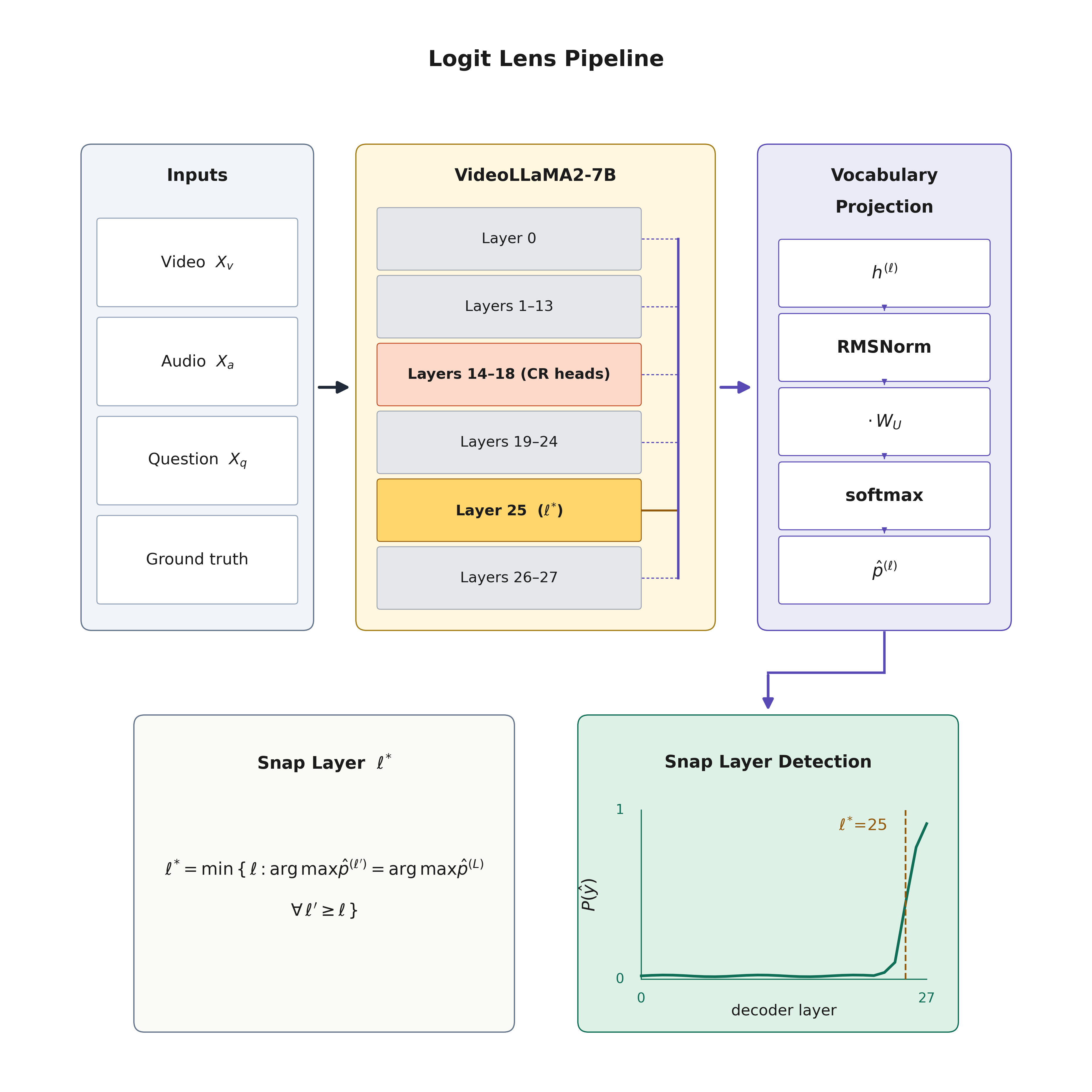}
\caption{The Logit Lens capture pipeline and snap layer detection. Residual streams are extracted across all layers to track the trajectory of the predicted target token.}
\label{fig:logit_lens_pipeline}
\end{figure}

\subsection{Mechanistic Audit}

\begin{figure}[t]
\centering
\includegraphics[width=\columnwidth]{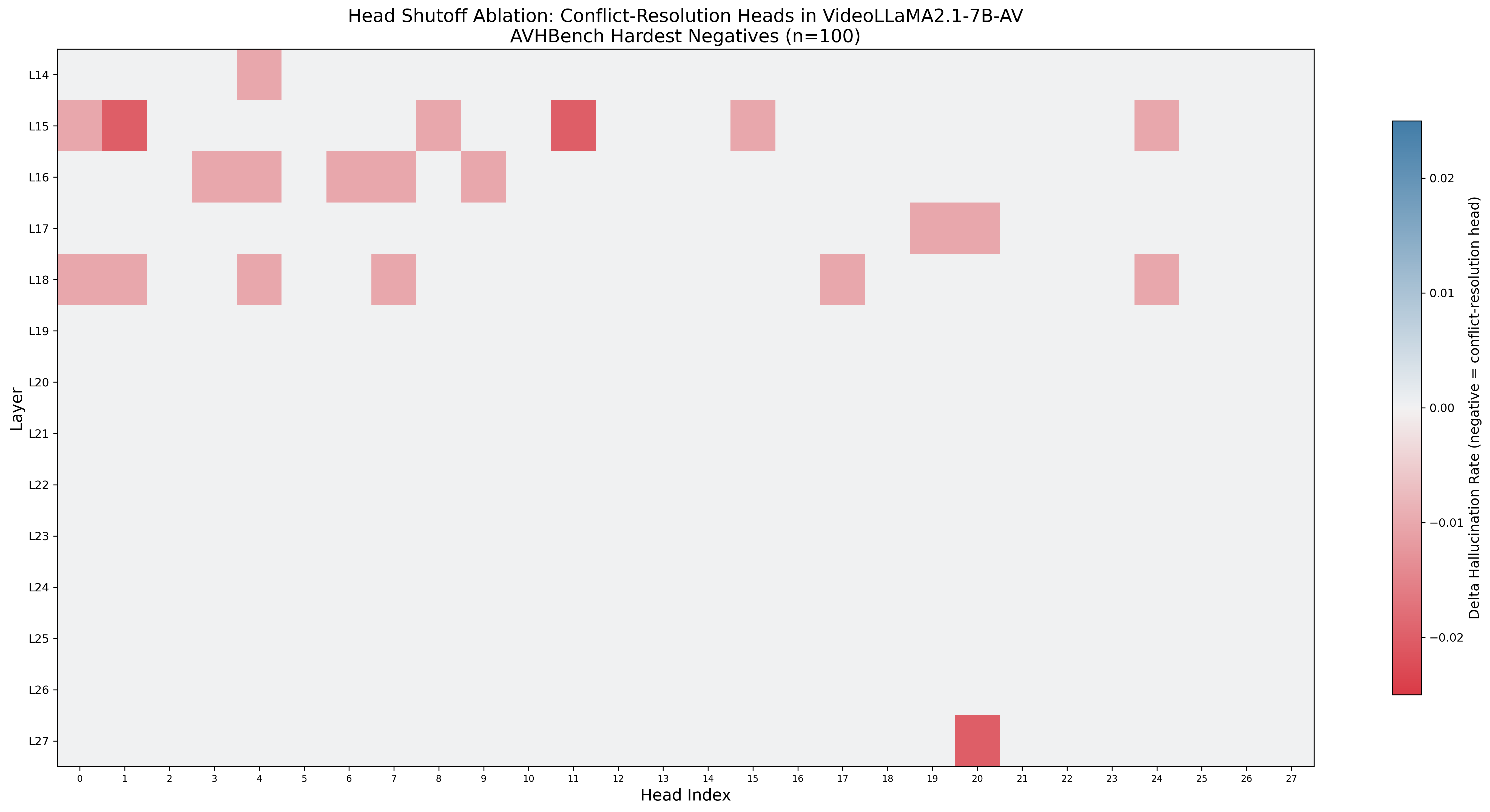}
\caption{Distribution of hallucination and conflict resolution
heads throughout layers 14–27 of VideoLLaMA2-7B. A positive
$\Delta$ (Blue) implies hallucination while a negative $\Delta$ (Red) represents
conflict-resolution}
\label{fig:mech_audit_CR}
\vspace{-8pt}
\end{figure}

We audit the baseline VideoLLaMA2-7B using 100 samples
from the conflict split. For each attention head in layers
14–27, we shut off that head (replacing its contribution with
its mean) and re-run inference. We define $\Delta$ as the change
in hallucination rate (fraction failing exact-match Yes/No)
relative to the unablated baseline: $\Delta \ge 0.01$ marks a hallucination head (ablation reduces hallucinations), $\Delta \le -0.01$
marks a conflict-resolution head (ablation increases them).
The threshold is operational, without statistical calibration.
For top candidates we follow up with activation patching
from clean to conflict samples to test generalization.

The audit identifies 21 conflict-resolution heads clustered
in layers 15–18; no heads pass the hallucination threshold. We do not interpret this absence as evidence of nonlocalizability: it can also reflect insufficient audit power,
distributed computation, or threshold sensitivity. The audit
was not run on the LoRA-tuned configurations.

\subsection{Logit Lens Capture Protocol}

We apply the logit lens across all layers of VideoLLaMA2-7B to locate the generative prior's emergence. For all $1{,}281$ conflict samples, we project the residual stream $h^{(\ell)}$ through the final RMSNorm and $W_U$ to yield a per-layer distribution $\hat{p}^{(\ell)}$. We track $P(\text{Yes})$, $P(\text{No})$, and the predicted token across layers, allowing us to define the snap layer $\ell^*$ where the top prediction stabilizes, revealing where the network commits to its prior.

\section{Results \& Discussion}
\subsection{Baseline Compositional Failure Under Conflict}
To establish the severity of this vulnerability prior to alignment interventions, we evaluated the off-the-shelf InternVideo2 on AVHBench. We find that while the model
achieves $60.1\%$ accuracy on the full AVHBench dataset,
performance drops to $27.8\%$ on the conflict set ($\Delta = 32.3\%$).
Since this is a binary Yes/No task, dropping below chance
level indicates the model is being actively misled by contradictory modalities, rather than randomly guessing. Further,
conflict triggers a $17.3\%$ instruction failure rate (1,108 out
of 6408 total inferences) where the model does not follow
the ”Yes/No” format, and generates its own filler. This confirms that under cross-modal conflict, the model abandons
composition for an internal generation prior, establishing
prior dominance as a systematic baseline vulnerability

\subsection{Bias shifts without compositional gains}

Table~\ref{results-table} shows two distinct patterns that should be read
separately: accuracy remains near chance across all configurations, while the Yes/No output prior shifts substantially.
ACTC is slightly Yes-leaning, TATI is more balanced, and
the full pipeline becomes strongly No-leaning. For context, the off-the-shelf VideoLLaMA 2-7B-AV base model
achieves 51.7$\%$ on the AV Matching task, showing that these
alignment interventions truly lower compositional conflict
resolution compared to pretrained baselines.

\begin{table}
  \caption{Exact-string Yes/No results on AVHBench. ``Scored $N$'' is the number of samples whose greedy completion is exactly \texttt{Yes} or \texttt{No}; remaining generations are omitted from this table. The main pattern is stable across rows: answer bias shifts, but accuracy stays near chance on the scored subset. All three methods failed to exceed \model{}'s base 51.7\% accuracy.}
  \label{results-table}
  \vskip 0.1in
  \begin{center}
    \begin{small}
      \begin{sc}
        \setlength{\tabcolsep}{3pt}
        \resizebox{\columnwidth}{!}{%
        \begin{tabular}{lccccc}
          \toprule
          Configuration & Scored $N$ & Overall Acc & GT=Yes Acc & GT=No Acc & Yes/No Preds \\
          \midrule
          ACTC & 5165 & 49.8\% & 55.5\% & 44.1\% & 2895 / 2270 \\
          ACTC + TATI & 5288 & 49.0\% & 46.9\% & 51.1\% & 2529 / 2759 \\
          ACTC + TATI + AMD & 5299 & 50.2\% & 35.1\% & 65.2\% & 1853 / 3446 \\
          InternVideo2 & 5302 & 52.3\% & 51.3\% & 53.3\% & 2600 / 2702 \\
          \bottomrule
        \end{tabular}}
      \end{sc}
    \end{small}
  \end{center}
  \vskip -0.1in
\end{table}

The full pipeline over-corrects toward \texttt{No}, with 3446 \texttt{No} predictions against 1853 \texttt{Yes} predictions, yet this bias shift does not yield better conflict resolution. Mapped onto the four-step framework from Section 2, this suggests failure specifically at the binding and compatibility judgment steps; the model changes which shortcut it prefers rather than
learning to compare modalities.

\subsection{Prior dominance appears in both captions and layer-wise traces}
Open-ended captioning exhibits the same pattern. Across
checkpoints, the model repeatedly begins with variants of
“A person is playing a musical instrument$\ldots$” (occurring in 42 of 59 open-ended captioning outputs), even when neither modality supports that claim. Because the surface continuation changes while the semantic scaffold remains stable, we interpret this as behavioral evidence for prior dominance: the model preserves fluency while losing grounding. Appendix D covers this correlation and possible interventions.

\begin{figure}
\centering
\includegraphics[width=\columnwidth]{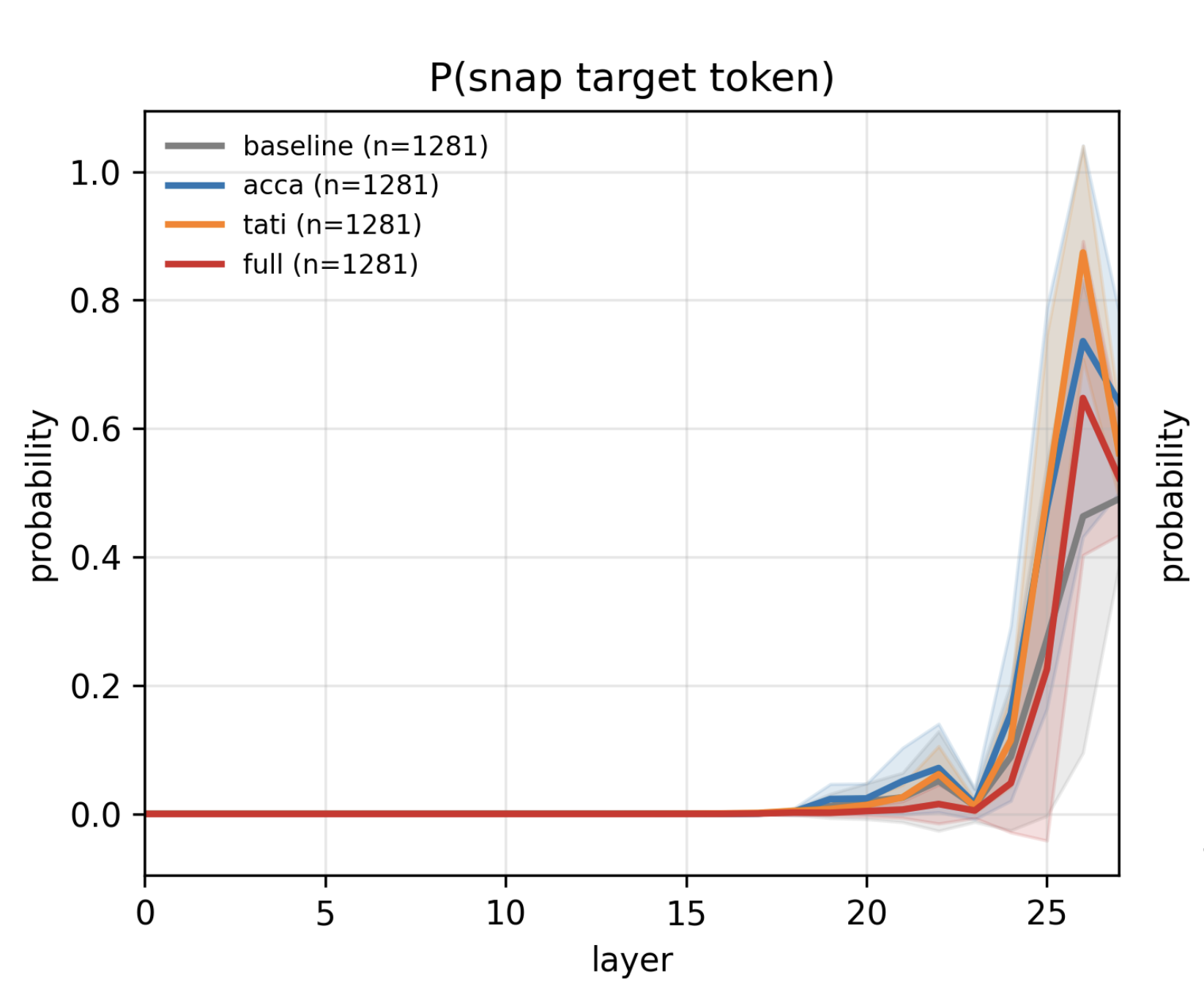}
\caption{Logit-lens probability trajectories for the target token. Across configurations, the prediction stabilizes around layer
$25.5 \pm 1$, indicating a consistent late-layer point at which the final
prediction becomes linearly decodable}
\label{fig:snap_layer_histogram}
\end{figure}

We find that all three configurations and the baseline model
of Video-LLaMA2-7B snap to their final-layer prediction $\ell^*$
at approximately layer 25.5 $\pm$1. Further, no configuration demonstrated significant target-token probability mass
before layer 18, meaning the prior forms well beyond the
model's conflict-resolution heads (which cluster between
layers 15 and 18). As the pipeline progresses, GT=Yes
accuracy collapses from $55.5\%$ to $35.1\%$, while GT=No
accuracy improves. This suggests the commitment at the
snap layer increasingly favors No, regardless of ground truth.
This gap suggests that conflict-relevant computation occurs
but is overridden downstream; the model detects the conflict
but does not act on it at generation time.

Logit lens analysis confirms that the shift in priors is mechanistic: each individual capture commits to either Yes/No
at the snap layer, matching the bias documented in Table~\ref{results-table}.
Figure \ref{fig:snap_layer_histogram} shows that the prior fires towards a fixed direction,
independent of the ground truth. Alignment fine-tuning
shifts which direction the prior fires, rather than when it
fires. After correcting for a temporal embedding artifact
by subtracting the shared $E_{\text{temp}}$ vector, AV survival remains
identical across all configurations (Appendix A, Figure 4).
The pipeline alters the surface-level bias, but deep semantic
routing remains structurally unchanged.

\section{Limitations} 
Our mechanistic evidence is limited to the VideoLLaMA 2
architecture, though our behavioral baseline includes InternVideo2. Table~\ref{results-table} is restricted to exact \texttt{Yes}/\texttt{No} completions
and should be read as a strict probe.

We note that the interleaving process increases the total
sequence length compared to the ACTC baseline. Our evaluation does not isolate the effects of increased sequence length from the temporal alignment itself, which remains a
limitation of the current TATI implementation.

\section{Conclusion} 
Under cross-modal conflict, neither off-the-shelf InternVideo2 nor aligned VideoLLaMA 2-7B-AV show strong
compositional generalization. VideoLLaMA 2-7B-AV does
not show improved conflict resolution after stronger alignment; instead, the interventions mainly reshape answer bias.

We find evidence consistent with prior dominance, specifically with a late-layer commitment to an internally preferred
response pattern, as a practical obstacle for compositional
audio-visual reasoning across both plain and aligned AVLLMs. In any embodied or multi-sensor setting, a model
that detects but overrides cross-modal conflict cannot be
trusted to act on contradictory evidence; making late-layer
commitment control a practical priority beyond benchmarks.
The immediate implication is methodological: future work
should evaluate conflict composition directly and pair alignment improvements with interventions that test, and ideally
control, late-layer commitment in safety-relevant settings.

\newpage
\appendix
\onecolumn

\section{Appendix}
\begin{figure}[h]
  \centering
  \includegraphics[width=0.9\textwidth,height=0.45\textheight,keepaspectratio]{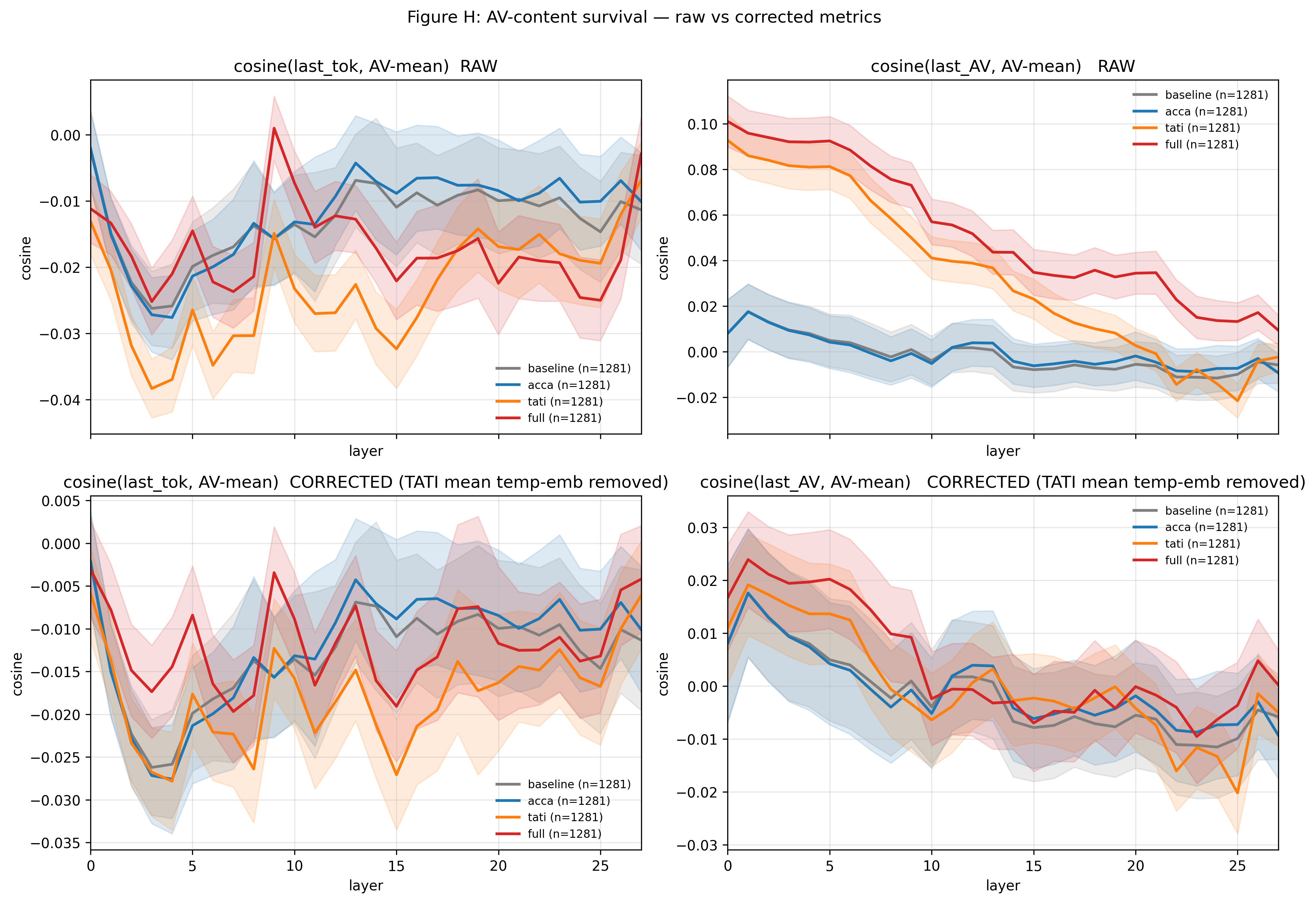}
  \caption{Exploratory cosine-similarity proxy $\avproxy$ across layers. We report this only as a heuristic representation diagnostic, not as a direct measure of modality attribution.}
  \label{fig:av_survival}
\end{figure}

Figure~\ref{fig:av_survival} plots an exploratory proxy, \(\avproxy\), defined as cosine similarity between late-layer audio and visual token states after removing the shared timestamp embedding used by TATI-style interleaving. We include this plot only as suggestive context. Because deep residual states have already mixed information through attention and MLP updates, this subtraction should not be interpreted as an exact decomposition of modality content. Accordingly, the main paper does not rely on this proxy for its core claim.

\section{Training Setup}
All three pipeline stages were implemented using Low-Rank Adaptation on top of the frozen \model{} backbone. Supervised fine-tuning was conducted on 8 A100 GPUs with a batch size of 64. Custom vision and audio projectors were trained alongside the LoRA adapters to map modality-specific features into the shared language-model embedding space.

\section{Expanded Related Works}
Recent benchmarks such as AVHBench \cite{sungbin2025avhbench}, AVCD \cite{jung2025avcd}, and broader multimodal hallucination evaluations \cite{leng2024cmm} establish the importance of testing grounded audio-visual reasoning. Temporal interleaving and synchronization are widely used design choices in AV-LLMs, including AVicuna-style interleaving \cite{tang2024avicuna}, while recent methods also target hallucination at training time or decoding time through stronger alignment or adaptive decoding \cite{guo2025dolphin,chung2026mad,jung2025forkmerge,chowdhury2025aurelia}. Our paper is complementary to that line of work: instead of proposing a new alignment module, it asks whether these alignment pressures improve composition specifically under contradiction.

\section{Discussion \& Future Mitigation Strategies}
While our mechanistic audit identifies a clear temporal gap between conflict-detection (layers 15--18) and prior commitment (layer 25.5), we acknowledge that this observation is currently correlational. It remains possible that the late-layer stabilization reflects downstream information propagation rather than a dedicated ``prior dominance'' mechanism. To distinguish between these hypotheses, future work should employ causal interventions,specifically, steering vectors or residual stream dampening,applied to layers 20--24. By dampening the model's internal representation of the ``prior'' immediately after the conflict-detection heads, one could test whether the model is forced to rely on earlier, grounded audio-visual computations. Additionally, while our audit uses an operational threshold of $\Delta = \pm0.01$, future work would benefit from statistical calibration through bootstrapping or randomized ablation baselines to rigorously quantify the impact of conflict-resolution heads. Finally, extending this analysis to structurally distinct architectures remains a priority to verify if the ``snap layer'' is a universal property of autoregressive multimodal systems.

\end{document}